\documentclass[runningheads]{llncs}

\usepackage{eccv}

\usepackage{eccvabbrv}

\usepackage{graphicx}
\usepackage{booktabs}

\usepackage[accsupp]{axessibility}  

\usepackage{hyperref}
\usepackage{custom}

\usepackage{orcidlink}

\usepackage{multirow}
\usepackage[table,xcdraw]{xcolor}
\usepackage[normalem]{ulem}
\useunder{\uline}{\ul}{}
\usepackage{wrapfig}
\usepackage{algorithm}
\usepackage{algorithmic}

\begin{document}

\title{ACT Now: Preempting LVLM Hallucinations \\via Adaptive Context Integration} 

\titlerunning{\hspace{1em}}

\author{Bei Yan\inst{1,2} \and
Yuecong Min\inst{1,2} \and
Jie Zhang\inst{1,2} \and
Shiguang Shan\inst{1,2} \and
Xilin Chen\inst{1,2}}

\authorrunning{\hspace{1em}}

\institute{State Key Laboratory of AI Safety, Institute of Computing Technology,\\Chinese Academy of Sciences, Beijing, China \and
University of Chinese Academy of Sciences, Beijing, China
}

\maketitle

\begin{abstract}
Large Vision-Language Models (LVLMs) frequently suffer from severe hallucination issues. Existing mitigation strategies predominantly rely on isolated, single-step states to enhance visual focus or suppress strong linguistic priors. However, these static approaches neglect dynamic context changes across the generation process and struggles to correct inherited information loss. To address this limitation, we propose \textbf{A}daptive \textbf{C}ontext in\textbf{T}egration (\textbf{ACT}), a training-free inference intervention method that mitigates hallucination through the adaptive integration of contextual information. Specifically, we first propose visual context exploration, which leverages spatio-temporal profiling to adaptively amplify attention heads responsible for visual exploration. To further facilitate vision-language alignment, we propose semantic context aggregation that marginalizes potential semantic queries to effectively aggregate visual evidence, thereby resolving the information loss caused by the discrete nature of token prediction. Extensive experiments across diverse LVLMs demonstrate that ACT significantly reduces hallucinations and achieves competitive results on both discriminative and generative benchmarks, acting as a robust and highly adaptable solution without compromising fundamental generation capabilities.
\end{abstract}

\section{Introduction}

Large Vision-Language Models (LVLMs) that integrate visual understanding with language generation have achieved remarkable progress in complex multi-modal tasks, such as image captioning and visual question answering~\cite{ccallava, chen2024internvl2}. However, LVLMs suffer from severe hallucination issues, frequently fabricating content that contradicts the visual input, such as non-existent objects or misidentifying attributes. This poses significant safety concerns, particularly as LVLMs are increasingly deployed in high-stakes, real-world applications~\cite{chen2026survey, sahoo2024survey}.

\begin{figure}[t]
\centering
\includegraphics[width=\linewidth]{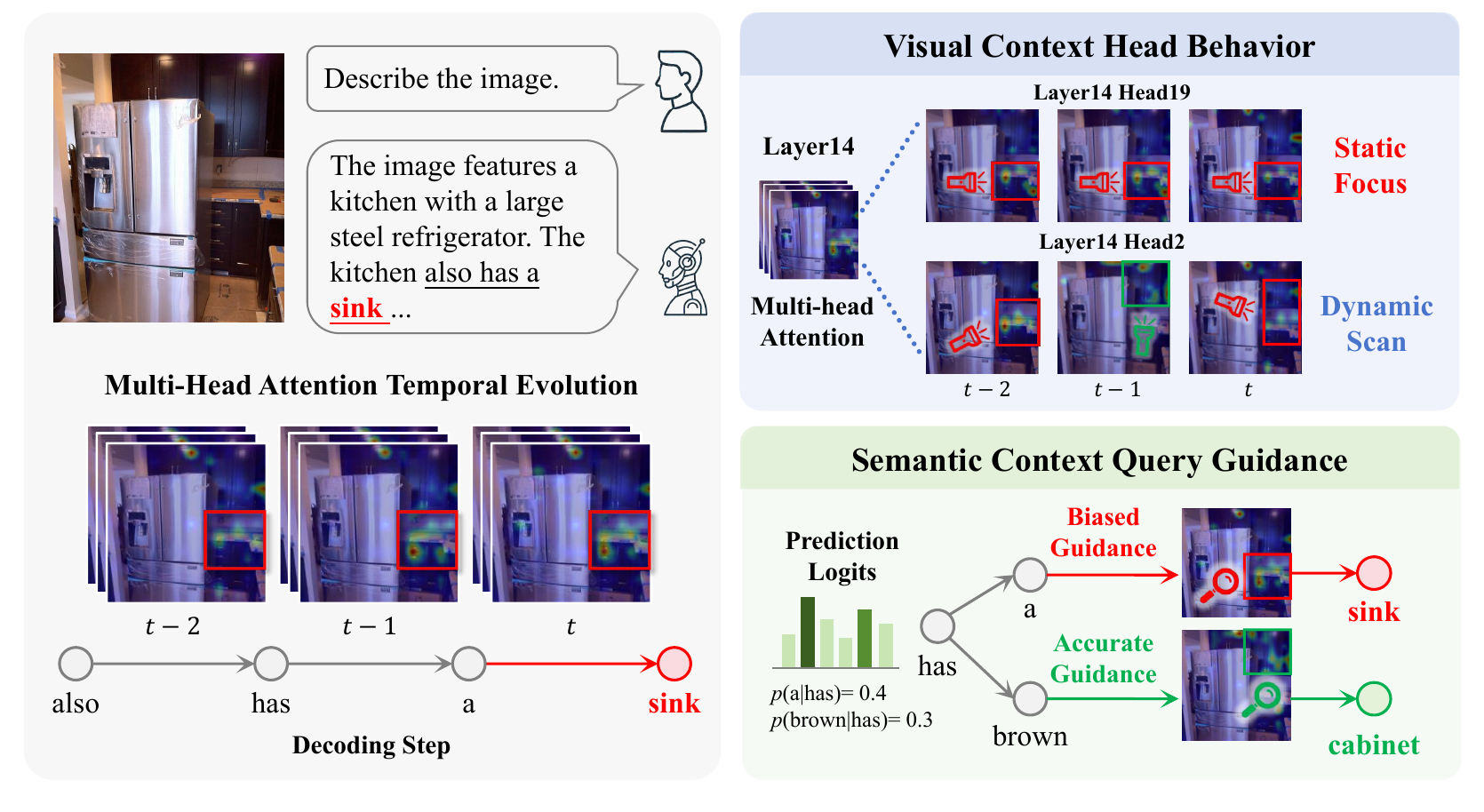}
\caption{Illustration of temporal evolution in cross-modal attention during generation. (Left) Visual attention typically surges before the model predicts the corresponding token. (Top Right) Attention heads show distinct behavioral patterns during decoding. (Bottom Right) Discretization of low-confidence predictions causes information loss, resulting in poor grounding and potential hallucinations.}
\label{fig:intro_motivation}
\end{figure}

To enhance the reliability of LVLMs, numerous studies~\cite{yin2024woodpecker,liu2024pai,leng2024vcd,huang2024opera,agla,VisAttnSink} have focused on mitigating hallucinations from either a visual or a linguistic perspective. Vision-centric methods typically attempt to explicitly or implicitly enhance visual focus on corresponding spatial regions through external grounding tools~\cite{yin2024woodpecker} or internal attention mechanisms~\cite{agla, liu2024pai, VisAttnSink}. On the other hand, language-based methods leverage advanced decoding strategies~\cite{huang2024opera} or contrastive tracing~\cite{leng2024vcd} to mitigate the model's over-reliance on strong linguistic priors. Although these methods have achieved promising results in hallucination reduction, they predominantly rely on isolated single-step states and neglect dynamic context changes across the generation process, potentially causing cascading errors and failing to correct inherited information loss during generation. As illustrated in the left panel of Fig.~\ref{fig:intro_motivation}, the visual attention allocated to a specific region often naturally surges just before the model explicitly predicts the corresponding token, underscoring the fundamental need to proactively integrate dynamic contextual information.

Driven by this insight, we analyze the temporal evolution of cross-modal attention. As shown in Fig.~\ref{fig:intro_motivation}, we observe two distinct behavioral patterns among attention heads during the generation process. Specifically, static heads persistently focus on fixed spatial regions across time steps, maintaining highly concentrated visual attention weights. In contrast, dynamic heads exhibit more distributed and lower-magnitude  attention, actively scanning and exploring new visual areas as the decoding sequence progresses. We hypothesize that a primary cause of hallucinations is that while dynamic heads may locate the correct visual evidence, their signals remain too weak to overcome the strong visual inertia of static heads. Furthermore, the uncertainty of auto-regressive decoding amplifies these sequential errors. Specifically, the discrete nature of token prediction discards the continuous semantic information required for precise alignment. In unclear scenes,this discretization loss yields insufficient grounding cues, causing attention queries to drift and ultimately triggering hallucinated predictions.

Inspired by these observations, we propose \textbf{A}daptive \textbf{C}ontext in\textbf{T}egration (\textbf{ACT}), a training-free inference intervention designed to preemptively calibrate attention allocation using two distinct contextual sources. Specifically, we first introduce Visual Context Exploration (VCE), which leverages spatio-temporal profiling to amplify the heads responsible for visual exploration, encouraging the model to gather sufficient evidence from broader spatial contexts. Moreover, we design Semantic Context Aggregation (SCA) to calculate a contextually fused query by marginalizing over parallel generation branches conditioned on different potential tokens, which effectively stabilizes the decoding process by reducing the risk of attention drift during highly uncertain steps. By employing the robust semantic queries from SCA to guide the targeted exploration in VCE, ACT significantly neutralizes the uncertainty inherent in local predictions and prevents the propagation of hallucinations.

To verify the effectiveness of our proposed method, we evaluate ACT across several state-of-the-art LVLM architectures using a diverse suite of discriminative and generative hallucination benchmarks~\cite{pope, wang2023amber, fumme, chair}. Experimental results demonstrate that ACT establishes a new state-of-the-art in hallucination mitigation. Crucially, as a training free approach that requires no external grounding tools, ACT achieves these improvements without compromising the fundamental multi-modal generation capabilities of LVLMs. Our main contributions are summarized as follows:

\begin{itemize}
    \item We propose the Adaptive Context Integration framework, which moves beyond traditional single-step mitigation strategies by proactively leveraging dynamic contextual information.
    \item We design Visual Context Exploration, which leverages spatio-temporal profiling to identify and amplify dynamic attention heads, significantly enhancing fine-grained visual perception.
    \item We introduce Semantic Context Aggregation, which marginalizes over parallel generation branches conditioned on different potential tokens, effectively preventing cascading semantic deviations.
\end{itemize}

\section{Related Work}
\subsection{Hallucination in LVLMs}
Research has increasingly focused on LVLM hallucinations, resulting in extensive evaluation benchmarks and diverse mitigation strategies~\cite{fumme, pope, wang2023amber, chair,liu2025phd}. 
Unlike resource-intensive fine-tuning mitigation approaches~\cite{ccallava}, training-free inference interventions have emerged as a highly efficient paradigm. One major category modifies the textual decoding distribution to suppress linguistic priors~\cite{woo2024ritual, zhuang2025vasparse}. For instance, DoLa~\cite{chuangdola} and VCD~\cite{leng2024vcd} contrast probability distributions across different layers or noised
inputs. Similarly, AGLA~\cite{agla} assembles output logits derived from global and locally masked image. Other strategies, such as HALC~\cite{halc}, DOPRA~\cite{wei2024dopra}, and ONLY~\cite{wan2025only}, adjust logits by integrating external visual constraints or penalizing predictive over-confidence. 

Another category involves more fundamental interventions by directly manipulating internal latent states~\cite{zou2025memvr}. Early works, such as PAI~\cite{liu2024pai} and VAF~\cite{yin2025vaf}, apply a global scaling to visual attention weights to compel a stronger visual focus. More recent studies target specific attention anomalies, for example, VAR~\cite{VisAttnSink} and TAME~\cite{tang2025tame} prevent attention from collapsing onto sink or anchor tokens. Despite their effectiveness, these methods mainly utilize instantaneous, single-step states, such as token logit distribution~\cite{leng2024vcd,agla}, or attention patterns~\cite{huang2024opera,VisAttnSink}, to perform interventions at the current decoding step. In contrast, our proposed ACT framework proactively integrates the continuous evolution of vision-language context to preemptively correct attention trajectories.
\subsection{Interpretability of LVLMs}
Recent advancements in mechanistic interpretability have sought to demystify the internal operations of LVLMs~\cite{kamath2023s,tong2024eyes,zheng2025lvlms}, particularly the interaction between visual and textual features. Several pioneering studies have analyzed these models from a structural perspective. At the layer level, research has revealed a functional hierarchy where middle layers typically perform visual enrichment, while deeper layers prioritize semantic refinement~\cite{jiang2025devils, kaduri2025whatinimage}. At a more granular head-wise level, analyses have identified specific attention heads with different functions, such as visual focusing and highly localized grounding~\cite{bi2025unveiling, groundinghead, zhang2024eah}. Despite these insights, existing studies predominantly characterize attention heads using static spatial patterns, overlooking their temporal evolution during autoregressive decoding. In contrast, our method explicitly profiles these temporal dynamics, distinguishing between static heads trapped in visual inertia and dynamic heads that actively explore evolving contexts.

\section{Method}

\begin{figure}[t]
\centering
\includegraphics[width=\linewidth]{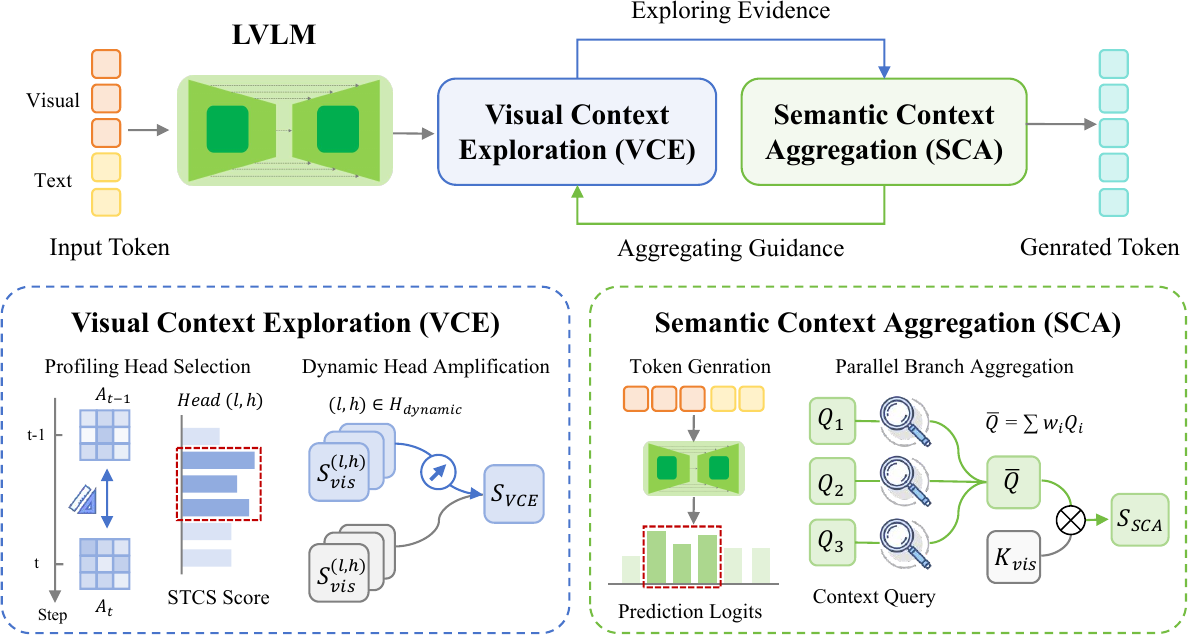}
\caption{Overview of the proposed ACT method. (Left) VCE amplifies dynamic heads to capture broader visual evidence. (Right) SCA marginalizes parallel textual hypotheses to preemptively resolve local linguistic uncertainty.}
\label{fig:overview}
\end{figure}

In this section, we first establish the preliminaries of the LVLM decoding process and the attention mechanism in Sect.~\ref{sec:preliminary}, then introduce the proposed ACT method. As illustrated in Fig.~\ref{fig:overview}, the core objective of ACT is to mitigate hallucination through the adaptive integration of contextual information. Specifically, we propose visual context exploration (VCE) to actively gather broader visual evidence for accurate grounding in Sect.~\ref{sec:vce} and semantic context aggregation (SCA) to preemptively resolve local linguistic uncertainty before errors cascade in Sect.~\ref{sec:sca}, thereby ensuring robust and reliable vision-language alignment.

\subsection{Preliminaries}
\label{sec:preliminary}
\textbf{Model Overview.} Recent LVLMs~\cite{ccallava, chen2024internvl2, Qwen2-VL} typically consist of three primary components: a vision encoder, a vision-language projector, and an Large Language Model (LLM) backbone. For a given image $v$ and text prompt $x$, the vision encoder and projector transform $v$ into a visual token sequence $X_{vis}$. This sequence is then concatenated with the text tokens $X_{text}$ to form a unified input for the LLM parameterized by $\theta$. The model then auto-regressively generates the target response sequence $y = \{y_1, y_2, \dots, y_T\}$, formulated as:
\begin{equation}
p(y \mid v, x; \theta)=\prod_{t=1}^{T} p(y_t \mid X_{vis}, X_{text}, y_{<t}; \theta),
\end{equation}
where $y_{<t}$ denotes the generated tokens before the current decoding step $t$.

\textbf{Attention Mechanism.} The LLM backbone is fundamentally built upon the Transformer architecture, which processes the input tokens into hidden states, with each layer incorporating a Multi-Head Attention (MHA) module followed by a Multi-Layer Perceptron (MLP). At a specific transformer layer $\ell$ and attention head $h$, the input hidden representations are projected into query, key, and value representations $Q^{(\ell, h)}$,  $K^{(\ell, h)}$, $V^{(\ell, h)} \in \mathbb{R}^{N \times d}$, where $N$ is the sequence length and $d$ is the head dimension. The attention weights $A^{(\ell, h)}$ are derived via the scaled dot-product and softmax function:
\begin{equation}
S^{(\ell, h)} = \frac{Q^{(\ell, h)} (K^{(\ell, h)})^\top}{\sqrt{d}}, \quad A^{(\ell, h)} = \text{Softmax}(S^{(\ell, h)}).
\end{equation}
To explicitly formulate our intervention, we decompose the pre-softmax attention scores $S^{(\ell, h)}$ and denote the segment indexed by the image tokens as $S^{(\ell, h)}_{vis}$. The proposed ACT method operates exclusively by modifying $S^{(\ell, h)}_{vis}$, which ensures the subsequent normalization naturally incorporates the updated visual scores without disrupting the fundamental language generation capabilities.

\subsection{Visual Context Exploration}
\label{sec:vce}
Extensive studies~\cite{groundinghead,jiang2025devils,bi2025unveiling} have investigated the internal visual processing mechanisms of LVLMs, particularly the attention patterns across different layers and heads. In contrast to existing works that predominantly focus on the attention behaviors within a single decoding step, we explore the evolution of visual attention across successive timesteps. 

To quantitatively identify these sequential behaviors, we introduce Spatio-Temporal Covariance Similarity (STCS) metric to measure the structural alignment of visual attention maps over a local temporal window. Let $\tilde{A}_{t}^{(\ell, h)} \in \mathbb{R}^{H \times W}$ denote the 2D spatial attention map produced by the $h$-th head at layer $\ell$ during decoding step $t$. To simplify notation, we use $A_t$ to represent $\tilde{A}_{t}^{(\ell, h)}$. We define $\rho(\cdot, \cdot)$ as the normalized centered Frobenius inner product to quantify the covariance alignment between two matrices. By modeling the spatial structural dependencies via row (\ie, $AA^\top$) and column (\ie, $A^\top A$) covariances, the instantaneous similarity $\phi(A_{t-1}, A_t)\in[0,1]$ between step $t-1$ and $t$ is concisely formulated as their mean alignment:
\begin{equation}
\phi(A_{t-1}, A_t) = \frac{1}{2} \Big( \rho\big(A_{t-1}A_{t-1}^\top, A_{t}A_{t}^\top\big) + \rho\big(A_{t-1}^\top A_{t-1}, A_{t}^\top A_{t}\big) \Big).
\end{equation}

To ensure robustness, we construct a small-scale calibration dataset (under 500 images) to compute the expected average similarity over a temporal window of size $\tau$. This profiling allows us to categorize attention heads based on their temporal stability:
\begin{itemize}
\item Dynamic Heads ($\mathcal{H}_{dynamic}$) are heads characterized by low average similarity scores, reflecting a continuous updating of visual representations to process varying regions of interest over time.
\item Static Heads ($\mathcal{H}_{static}$) are heads with scores approaching 1, indicating a tendency to focus consistently on invariant spatial regions across steps.
\end{itemize}

During inference, we specifically target the middle layers of the LVLM, denoted as $\ell \in [\ell_s, \ell_e]$, to encourage the model in adaptively exploring the visual context. As indicated by previous research~\cite{jiang2025devils, kaduri2025whatinimage}, these layers feature deep cross-modal interactions and are primarily responsible for visual information enrichment. Formally, for each identified dynamic head $h \in \mathcal{H}_{dynamic}$ in layer $\ell \in [\ell_s, \ell_e]$, we amplify the visual attention scores by:
\begin{equation}
S_{VCE}^{(\ell, h)} = S_{vis}^{(\ell, h)} + \alpha \big| S_{vis}^{(\ell, h)} \big|,
\end{equation}
where $\alpha>0$ is a scaling factor. 

Note that the shallow layers $\ell \in [0, \ell_s]$ are primarily responsible for establishing global visual understanding~\cite{jiang2025devils, kaduri2025whatinimage, zhang2024eah}. To effectively consolidate this global contextual foundation during the prefill stage, we apply a consistent enhancement to each static head $h \in \mathcal{H}_{static}$. This stage-specific guidance ensures that while the model builds a robust global representation in the early stages, it maintains high sensitivity to localized visual nuances during deeper processing.

\subsection{Semantic Context Aggregation}
\label{sec:sca}
While the VCE module encourages active vision context exploration, visual hallucinations can also originate from the language side. Specifically, the discrete nature of token prediction introduces uncertainty and cumulative information loss during the decoding process~\cite{peng2024transofmerlimitations}. When encountering ambiguous contexts, a deterministic single token selection strategy often prematurely prunes alternative valid contexts. These potentially biased textual tokens then formulate a skewed query in subsequent steps, triggering a feedback loop that retrieves misaligned visual evidence and ultimately amplifies cascading hallucinations.

To address this limitation, we propose a semantic context aggregation method to maintain multiple contextual hypotheses. Specifically, we retain the Top-$K$ parallel generation branches, each conditioned on a distinct potential token from the previous step $t-1$. These branches are associated with normalized probability weights $w_i \in [0, 1]$, where $\sum_{i=1}^K w_i = 1$. At the current decoding step $t$, these $K$ parallel textual contexts are independently projected into $K$ distinct attention queries for a specific head $h$ at layer $\ell$, denoted as $\{Q_i^{( \ell, h)}\}_{i=1}^K$. By utilizing this expanded set of queries, the model can effectively aggregate visual evidence across diverse semantic interpretations.

Instead of allowing the most probable branch to process visual tokens independently, which may lead to divergent and hallucinated generation trajectories, we compute a contextually fused query by simply marginalizing over the parallel textual branches:
\begin{equation}
\bar{Q}^{(\ell, h)} = \sum_{i=1}^K w_i Q_i^{(\ell, h)}.
\end{equation}
We then compute the expected visual attention scores $S_{SCA}^{(\ell, h)}$ by performing the scaled dot-product between the fused query and the shared visual keys $K_{vis}^{(\ell, h)}$:
\begin{equation}
S_{SCA}^{(\ell, h)} = \frac{\bar{Q}^{(\ell, h)} (K_{vis}^{(\ell, h)})^\top}{\sqrt{d}}.
\end{equation}
This probabilistically fused attention score is then broadcasted to all $K$ branches to serve as the shared, consensus visual attention logit. By leveraging the expectation over diverse textual contexts, SCA effectively smooths the attention variance induced by local token uncertainty, providing a more robust visual query formulation.

\subsection{Adaptive Context Integration}

To fully harness the complementary strengths of vision-side context exploration and language-side context aggregation, we unify the VCE and SCA modules into the cohesive ACT method. Specifically, rather than independently amplifying the raw visual attention, we leverage the robust, fused visual attention scores $\bar{S}_{vis}^{(\ell, h)}$ derived from SCA to replace the head amplification in VCE. 

Formally, for each targeted head $h$ at layer $\ell$, the unified update rule for the visual attention scores is formulated as:
\begin{equation}
S_{ACT}^{(\ell, h)} = S_{vis}^{(\ell, h)} + \alpha \big| S_{SCA}^{(\ell, h)} \big|.
\end{equation}
By modulating the original attention with this probabilistically fused prior, ACT promotes a more reliable and nuanced visual-textual alignment during the inference process.

The updated visual scores $\hat{S}_{vis}^{(\ell, h)}\in\{S_{VCE}^{(\ell, h)}, S_{SCA}^{(\ell, h)}, S_{ACT}^{(\ell, h)}\}$ can be subsequently concatenated with the original textual scores $S_{text}^{(\ell, h)}$ along the sequence dimension to obtain the final attention $\hat{A}^{(\ell, h)}$ via the Softmax operation:
\begin{equation}
\hat{A}^{(\ell, h)} = \text{Softmax}\Big(\big[\hat{S}_{vis}^{(\ell, h)}, S_{text}^{(\ell, h)}\big]\Big).
\end{equation}

Through this adaptive integration, ACT proactively aligns multi-modal contexts, effectively smoothing the attention variance induced by linguistic uncertainty while simultaneously encouraging fine-grained visual exploration. This synergy significantly mitigates cascading visual hallucinations without compromising the fundamental generation capabilities of the LLM backbone.

\section{Experiments}

\subsection{Experimental Setup}

\textbf{Models.} To demonstrate the broad applicability and robustness of our proposed method across diverse architectures, we evaluate ACT on five state-of-the-art open-source LVLMs, including LLaVA-1.5~\cite{liu2023llava}, InternVL2~\cite{chen2024internvl2}, InternVL2.5~\cite{chen2024internvl25}, Qwen2-VL~\cite{Qwen2-VL}, and Qwen2.5-VL~\cite{Qwen2.5-VL}.

\textbf{Evaluation Benchmarks.} To rigorously assess the effectiveness of ACT in mitigating object hallucinations, we conduct a comprehensive evaluation across both discriminative and generative benchmarks. For discriminative tasks, we utilize POPE~\cite{pope}, the hallucination subset of MME~\cite{fumme}, and the discriminative subset of AMBER~\cite{wang2023amber} (denoted as AMBER-d). For generative tasks, we utilize CHAIR~\cite{chair} and the generative subset of AMBER~\cite{wang2023amber} (denoted as AMBER-g), setting the maximum number of generated tokens to 512. All evaluations adhere to the original protocols and metrics of each benchmark.

\textbf{Implementation Details.} For the VCE module, we construct the offline calibration set by randomly sampling 500 images from the MSCOCO~\cite{mscoco} dataset to compute the STCS metric. For LLaVA-1.5-7B, we select the $16$ heads with the lowest STCS scores in each layer based on the profiled scores to form the dynamic head set $\mathcal{H}_{dynamic}$, while the remaining heads constitute the static head set $\mathcal{H}_{static}$. The scaling factor is set to $\alpha = 0.6$, and the layer range $[\ell_s, \ell_e]$ is set to $[10, 26]$. For the SCA module, the number of parallel contextual generation branches is set to $K = 5$. More implementation details for all evaluated models are referred to Appendix. All experiments are conducted under zero-shot settings without any additional training.

\subsection{Main Results}

\begin{table}[t]
\centering
\caption{Comparison with state-of-the-art hallucination mitigation methods on LLaVA-1.5-7B. The top-2 results are \textbf{bolded} and {\ul underlined}, respectively.}
\label{tab:sota_comparison}
\setlength{\tabcolsep}{4pt}
\resizebox{\textwidth}{!}{%
\begin{tabular}{l|cc|ccccc|ccc}
\toprule
\multicolumn{1}{c|}{}& \multicolumn{2}{c|}{\textbf{POPE}}& \multicolumn{5}{c|}{\textbf{MME}} & \multicolumn{3}{c}{\textbf{CHAIR}}\\
\multicolumn{1}{l|}{\multirow{-2}{*}{{\textbf{Method}}}} & Acc↑ & F1↑ & Exist.↑ & Count↑ & Pos.↑ & Color↑ & Total↑ & $\text{C}_S$↓ & $\text{C}_I$↓ & Length   \\

\midrule
Dola~\cite{chuangdola}& 83.1 & 80.2& 180.1   & 127.4  & 119.3 & 154.6  & 594.1   & 57.0& 15.2& 97.5  \\
VCD~\cite{leng2024vcd}& 85.0 & 85.3& 184.7   & 137.3  & 128.7 & 153.0  & 603.7   & 51.0& 14.9& 101.9 \\
OPERA~\cite{huang2024opera}& 85.2 & 84.2& 180.7   & 133.3  & 111.7 & 123.3  & 549.0   & 47.0& 14.6& 95.3  \\
DOPRA~\cite{wei2024dopra} & 84.3 & 84.6& 185.7   & 138.3  & 120.7 & 133.0  & 577.7   & 46.3& 13.8& 96.1  \\
HALC~\cite{halc}& 84.0 & 83.9& 190.0   & 143.3  & 128.3 & 160.0  & 621.6   & 50.2& 12.4& 97.2  \\
CCA~\cite{ccallava}& 86.5  & 86.4 & 190.0   & 148.3  & 128.3 & 153.0  & 641.7   & 43.0& 11.5& 96.6  \\
RITUAL~\cite{woo2024ritual} & 84.3 & 85.2& 187.5   & 139.6  & 125.0 & 164.2  & 616.3   & 45.2& 13.2& 99.2  \\
EAH~\cite{zhang2024eah} & 86.0 & 85.7& 190.0   & 108.3  & {\ul 145.0}  & 160.7  & 604.0   & {\ul 36.4}& {\ul 9.9} & 97.7  \\
SID~\cite{SID}& 85.8 & 85.6& 183.9   & 132.2  & 127.8 & 155.9  & 599.8   & 44.2& 12.2& 99.4  \\
TAME~\cite{tang2025tame} & 85.7 & 85.4& {\ul 193.0}   & 137.3  & 139.0 & 164.7  & 634.0   & 41.3& 12.2& 98.8  \\
VAR~\cite{VisAttnSink} & {86.5}  & 86.0& 190.0   & 148.3  & 138.3 & 155.0  & 631.3   & 52.4& 14.5& 103   \\
AGLA~\cite{agla}& 85.5 & 84.6& \textbf{195.0} & {\ul 153.9}  & 129.4 & 161.7  & {640.0}& 43.0& 14.1& 98.8  \\
ONLY~\cite{wan2025only}& 85.1 & 85.5& 191.7   & 145.6  & 136.7 & 161.7  & 635.6   & 49.8& 14.3& 99.7  \\
MEMVR~\cite{zou2025memvr} &{\ul 87.4} &	{\ul 87.1}&190.0 	&\textbf{155.0} &	133.3 	&\textbf{170.6} &	{\ul 648.3}& 	46.6 	&13.0 &	99.6 \\
\rowcolor[HTML]{E7ECF8} 
\textbf{ACT(Ours)}   & \textbf{88.0}   & \textbf{87.7}  & \textbf{195.0} & 153.3   & \textbf{150.0}   & {\ul 170.0}   & \textbf{668.3} & \textbf{33.0}  & \textbf{9.3}   & 92.4  \\ 
\bottomrule
\end{tabular}
}
\end{table}
\textbf{Comparison with State-of-the-Art Methods.} 
We first evaluate the effectiveness of our proposed ACT against a wide range of state-of-the-art hallucination mitigation methods. For a fair comparison, all methods in this setting are applied to the widely used LLaVA-1.5-7B backbone. The quantitative results are summarized in Table~\ref{tab:sota_comparison}.

ACT demonstrates superior performance across both discriminative and generative benchmarks. On the POPE benchmark, ACT achieves the highest Accuracy 88.0\% and F1 score $87.7\%$, outperforming decoding-based methods such as AGLA~\cite{agla}, SFT-based approaches like CCA~\cite{ccallava}, and attention-based methods including VAR~\cite{VisAttnSink}. Notably, these gains are obtained without additional tool usage or extra training cost, highlighting the efficiency and effectiveness of ACT. Furthermore, on the MME Hallucination subset, ACT establishes a total score of 668.3, exhibiting particularly strong gains in the Position and Color sub-tasks, which indicates that the proposed ACT significantly enhances the model's ability to accurately ground fine-grained visual attributes. 

In the generative CHAIR evaluation, ACT achieves the lowest object hallucination rates, significantly reducing both $\text{CHAIR}_S$ ($\text{C}_S$) to 33.0 and $\text{CHAIR}_I$ ($\text{C}_I$) to 9.3. While we observe a slight decrease in generation length
as a common trade-off in hallucination mitigation where models become more cautious and concise, the substantial drop in hallucination metrics confirms that ACT successfully prevents the model from fabricating non-existent objects, producing much safer and more faithful descriptions.

\textbf{Broad Applicability Across Diverse LVLMs.} 
A key advantage of our ACT framework is its training-free, plug-and-play nature. To demonstrate its broad applicability, we integrate ACT into five different state-of-the-art LVLMs, including LLaVA-1.5-7B/13B, InternVL2-8B, InternVL2.5-8B, Qwen2-VL-7B, and Qwen2.5-VL-7B. Table~\ref{tab:generalizability} details the results across both discriminative and generative benchmarks, ACT consistently enhances the baseline performance across almost all tested architectures without requiring any model-specific tuning. For instance, whether applied to the LLaVA architecture or the more recent Qwen2.5-VL and InternVL2.5 models, the addition of ACT (+ACT) consistently reduces generative hallucination metrics while maintaining or improving discriminative accuracy, which addresses a fundamental mechanism of visual hallucination prevalent across different transformer-based LVLM architectures.

\begin{table}[t]
\centering
\caption{Evaluation results of the proposed ACT across different LVLM architectures.}
\label{tab:generalizability}
\resizebox{\textwidth}{!}{%
\setlength{\tabcolsep}{2pt}
\begin{tabular}{l|c|c|c|ccc|ccc}
\toprule
& \textbf{POPE} & \textbf{MME}   & \textbf{AMBER-d} & \multicolumn{3}{c|}{\textbf{AMBER-g}}   & \multicolumn{3}{c}{\textbf{CHAIR}}\\
\multicolumn{1}{l|}{\multirow{-2}{*}{\textbf{Method}}}  & Acc↑ & Total↑ & Acc↑ & CHAIR↓ & Hal↓ & Cog↓ & $\text{C}_S$↓ & $\text{C}_I$↓ & F1↑ \\
\midrule
LLaVA-1.5-7B   & 85.2 & 638.3 & 71.6  & 7.5 & 35.9   & 4.3 & 51.4   & 15.2   & 75.8\\
\rowcolor[HTML]{E7ECF8}
+ACT & 88.0\tiny{($\uparrow$2.8)} & 668.3\tiny{($\uparrow$30.0)} & 77.0\tiny{($\uparrow$5.4)}  & 5.1\tiny{($\downarrow$2.4)} & 27.4\tiny{($\downarrow$8.5)}   & 2.7\tiny{($\downarrow$1.6)} & 33.0\tiny{($\downarrow$18.4)}   & 9.3\tiny{($\downarrow$5.9)} & 78.3\tiny{($\uparrow$2.5)}\\
LLaVA-1.5-13B  & 83.9 & 636.7 & 68.9  & 7.2 & 34.2   & 3.8 & 52.6   & 13.9   & 77.5\\
\rowcolor[HTML]{E7ECF8}
+ACT & 86.6\tiny{($\uparrow$2.7)} & 645.0\tiny{($\uparrow$8.3)} & 74.0\tiny{($\uparrow$5.1)}  & 4.6\tiny{($\downarrow$2.6)} & 25.3\tiny{($\downarrow$8.9)}   & 2.3\tiny{($\downarrow$1.5)} & 40.8\tiny{($\downarrow$11.8)}   & 10.7\tiny{($\downarrow$3.2)} & 79.1\tiny{($\uparrow$1.6)}\\
InternVL2-8B   & 85.6 & 701.7 & 71.2  & 8.3 & 68.2   & 8.1 & 49.6   & 10.3   & 73.9\\
\rowcolor[HTML]{E7ECF8}
+ACT & 88.5\tiny{($\uparrow$2.9)} & 711.7\tiny{($\uparrow$10.0)} & 81.8\tiny{($\uparrow$10.6)}  & 7.6\tiny{($\downarrow$0.7)} & 56.2\tiny{($\downarrow$12.0)}   & 6.6\tiny{($\downarrow$1.5)} & 41.6\tiny{($\downarrow$8.0)}   & 10.1\tiny{($\downarrow$0.2)} & 73.9\tiny{($-$0.0)}\\
InternVL2.5-8B & 88.9 & 705.0 & 83.4  & 7.6 & 63.5   & 7.3 & 35.8   & 9.1 & 76.1\\
\rowcolor[HTML]{E7ECF8}
+ACT & 89.4\tiny{($\uparrow$0.5)} & 710.0\tiny{($\uparrow$5.0)} & 84.8\tiny{($\uparrow$1.4)}  & 7.1\tiny{($\downarrow$0.5)} & 63.8\tiny{($\uparrow$0.3)}   & 7.6\tiny{($\uparrow$0.3)} & 32.2\tiny{($\downarrow$3.6)}   & 8.9\tiny{($\downarrow$0.2)} & 75.9\tiny{($\downarrow$0.2)}\\
Qwen2VL-7B & 88.4 & 688.3 & 86.6  & 6.5 & 42.8   & 4.0 & 27.2   & 9.3 & 76.2\\
\rowcolor[HTML]{E7ECF8}
+ACT & 89.1\tiny{($\uparrow$0.7)} & 693.3\tiny{($\uparrow$5.0)} & 89.2\tiny{($\uparrow$2.6)}  & 4.9\tiny{($\downarrow$1.6)} & 20.4\tiny{($\downarrow$22.4)}   & 1.2\tiny{($\downarrow$2.8)} & 21.2\tiny{($\downarrow$6.0)}   & 7.5\tiny{($\downarrow$1.8)} & 71.6\tiny{($\downarrow$4.6)}\\
Qwen2.5VL-7B   & 82.1 & 700.0 & 77.8  & 5.4 & 22.6   & 1.0 & 29.8   & 10.6   & 68.4\\
\rowcolor[HTML]{E7ECF8}
+ACT & 84.1\tiny{($\uparrow$2.0)} & 705.0\tiny{($\uparrow$5.0)} & 78.9\tiny{($\uparrow$1.1)}  & 4.5\tiny{($\downarrow$0.9)} & 21.8\tiny{($\downarrow$0.8)}   & 1.3\tiny{($\uparrow$0.3)} & 30.8\tiny{($\uparrow$1.0)}   & 9.4\tiny{($\downarrow$1.2)} & 69.2\tiny{($\uparrow$0.8)}\\

\bottomrule
\end{tabular}
}
\end{table}

\subsection{Ablation Studies}
\label{sec:ablation}

We perform comprehensive ablation studies on the LLaVA-1.5-7B model to isolate the contributions of our proposed approach. This includes a detailed evaluation of individual modules, head selection criteria, and profiling metrics.

\begin{table}[t]
\centering
\caption{Ablation of core modules in ACT.}
\label{tab:ablation_components}
\resizebox{\textwidth}{!}{%
\setlength{\tabcolsep}{4pt}
\begin{tabular}{l|c|c|c|ccc|ccc}
\toprule
& \textbf{POPE}& \textbf{MME}   & \textbf{AMBER-d} & \multicolumn{3}{c|}{\textbf{AMBER-g}}   & \multicolumn{3}{c}{\textbf{CHAIR}}\\
\multicolumn{1}{l|}{\multirow{-2}{*}{{\textbf{Method}}}}  & Acc↑ & Total↑ & Acc↑ & CHAIR↓ & Hal↓ & Cog↓ & $\text{CHAIR}_S$↓ & $\text{CHAIR}_I$↓ & F1↑ \\
\midrule
Baseline  & 85.2& 638.3& 71.6& 7.5 & 35.9 & 4.3 & 51.4 & 15.2 & 75.8 \\
+ SCA&85.2& 638.3& 71.6 & 7.5 & 33.1 & 4.1 & 46.8 & 13.8 & 76.5 \\
+ VCE& \textbf{88.0}& \textbf{668.3}& \textbf{77.0}& {\ul 5.3} & \textbf{27.2} & {\ul 2.8} & {\ul 35.4} & {\ul 9.6}  & {\ul 78.0} \\
\rowcolor[HTML]{E7ECF8} 
+ ACT& \textbf{88.0} & \textbf{668.3}& \textbf{77.0} & \textbf{5.1} & {\ul 27.4} & \textbf{2.7} & \textbf{33.0} &\textbf{9.3}  & \textbf{78.3}
\\ 
\bottomrule
\end{tabular}
}
\end{table}

\textbf{Effectiveness of Core Modules.}
We first dissect the overall ACT method to evaluate the individual contributions of the VCE and SCA modules. As shown in Table~\ref{tab:ablation_components}, compared to the baseline, the standalone SCA module effectively reduces hallucinations in generative tasks, \eg, lowering CHAIRs from 51.4 to 46.8. Since SCA aggregates information based on parallel textual contexts, it relies on previously generated tokens. Consequently, it primarily benefits continuous generation tasks without altering discriminative performance. Conversely, the VCE module intervenes directly at the decoder layer,
yielding substantial improvements across both discriminative and generative benchmarks. The combination of these two modules exhibit strong synergy, achieving the lowest hallucination rates and the highest overall accuracy, confirming the necessity of aligning vision and language contexts simultaneously.

\begin{table}[t]
\centering
\caption{Comparison of different head profiling metrics in ACT.}
\label{tab:ablation_metrics}
\resizebox{\textwidth}{!}{%
\setlength{\tabcolsep}{4pt}
\begin{tabular}{l|c|c|c|ccc|ccc}
\toprule
& \textbf{POPE}& \textbf{MME}   & \textbf{AMBER-d} & \multicolumn{3}{c|}{\textbf{AMBER-g}}   & \multicolumn{3}{c}{\textbf{CHAIR}}\\
\multicolumn{1}{l|}{\multirow{-2}{*}{{\textbf{Metric}}}}  & Acc↑ & Total↑ & Acc↑ & CHAIR↓ & Hal↓ & Cog↓ & $\text{CHAIR}_S$↓ & $\text{CHAIR}_I$↓ & F1↑ \\
\midrule
Entropy Change & 85.3& 618.3& 73.1& 5.9& 29.1& 3.0& {\ul 41.8}& {\ul 11.7}   & {\ul 77.3}\\
JSD  & 87.1& 570.0& 75.5& 5.4& 27.5& {\ul 2.5}& 46.8& 12.8 & 77.1\\
SSIM & {\ul 87.6}& 593.3& 75.7& \textbf{5.0} & {\ul 26.1}& \textbf{2.3} & 46.6& 13.0 & 77.0\\
CoM Shift  & 87.3& {\ul 635.0}& {\ul 76.2}& 6.0& 29.1& 2.8& 51.0& 14.2 & 76.1\\
\rowcolor[HTML]{E7ECF8} 
STCS(Ours)  & \textbf{88.0} & \textbf{668.3} & \textbf{77.0} & {\ul 5.1}& \textbf{27.4} & 2.7& \textbf{33.0} & \textbf{9.3} & \textbf{78.3}
\\ 
\bottomrule
\end{tabular}
}
\end{table}

\begin{table}[t]
\centering
\begin{minipage}[t]{0.405\textwidth}
\centering
\caption{Comparison of head selection strategy in VCE.}
\label{tab:ablation_scope}
\resizebox{\textwidth}{!}{%
\setlength{\tabcolsep}{4pt}
\begin{tabular}{l|c|c|c}
\toprule
& \textbf{POPE}& \textbf{MME}   & \textbf{AMBER-d} \\
\multicolumn{1}{l|}{\multirow{-2}{*}{{\textbf{Strategy}}}}  & Acc↑ & Total↑ & Acc↑ \\
\midrule
Baseline& 85.2 &  {\ul 638.3}& 71.6\\
Global& {\ul 87.3} & 630.0& {\ul 75.9}\\
\rowcolor[HTML]{E7ECF8} 
VCE&\textbf{88.0}& \textbf{668.3}   & \textbf{77.0}  
\\ 
\bottomrule
\end{tabular}
}
\end{minipage}\hfill
\begin{minipage}[t]{0.58\textwidth}
\centering
\caption{Comparison of SCA with other decoding strategies on CHAIR.}
\label{tab:ablation_decoding}
\resizebox{\textwidth}{!}{%
\setlength{\tabcolsep}{2pt}
\begin{tabular}{l|ccccc}
\toprule
\textbf{Strategy} & \textbf{$\text{CHAIR}_S$↓} &\textbf{$\text{CHAIR}_I$↓}& \textbf{Recall↑}& \textbf{Precision↑}& \textbf{F1↑} \\
\midrule
Greedy  & {\ul 51.4}& {\ul 15.2} & {\ul 78.1}& 73.6 & 75.8 \\
Nucleus & 58.6 & 18.7 & 75.1 & 67.8 & 71.3 \\
Beam  & 55.2 & 15.5 & \textbf{79.6} & {\ul 73.7} & \textbf{76.5} \\
\rowcolor[HTML]{E7ECF8} 
SCA & \textbf{46.8} & \textbf{13.8} & 77.3 & \textbf{75.7} & \textbf{76.5}\\ 
\bottomrule
\end{tabular}
}
\end{minipage}
\end{table}

\textbf{Efficacy of the STCS Metric.}
To identify the dynamic heads for exploration, ACT relies on STCS metric to quantify the sequential variations in visual attention distributions. To evaluate this choice, we replace STCS with alternative distribution discrepancy metrics, including Entropy Change, Jensen-Shannon Divergence (JSD), Structural Similarity Index (SSIM), and Center of Mass (CoM) Shift. As shown in Table~\ref{tab:ablation_metrics}, while alternative metrics provide some degree of hallucination mitigation, the STCS-guided ACT significantly outperforms them. We hypothesize that this disparity arises because traditional statistical metrics fail to fully capture 2D spatial dependencies within visual attention maps. For instance, Entropy Change merely measures the overall dispersion of attention weights, failing to account for spatial shifts in the distribution center. Attention shifting sharply from one object to another might yield zero entropy change. Similarly, point-wise metrics like JSD or coordinate-based metrics like CoM Shift struggle to capture holistic structural deformations. In contrast, STCS explicitly models the structural evolution of attention via spatial covariance, providing a much more accurate proxy for identifying dynamic heads and guiding the final multi-modal context integration. Detailed metric comparisons are provided in the supplementary material.

\textbf{Targeted vs. Global Head Enhancement.}
In the VCE module, we explicitly identify and amplify dynamic heads. To validate this targeted selection strategy, we compare it against a naive method that amplifies all attention heads uniformly, denoted as Global. As shown in Table~\ref{tab:ablation_scope}, although enhancing all heads without selection can reduce hallucination on POPE and AMBER, it degrades general multi-modal capabilities on MME from 638.3 to 630.0. This degradation likely occurs because uniformly amplifying an inherently misaligned attention distribution introduces noise. By strictly targeting the dynamic exploration heads, our method safely boosts proactive visual grounding without compromising the model's structural integrity.

\begin{figure}[t]
    \centering
        \begin{minipage}{0.33\linewidth}
        \centering
        \includegraphics[width=\linewidth]{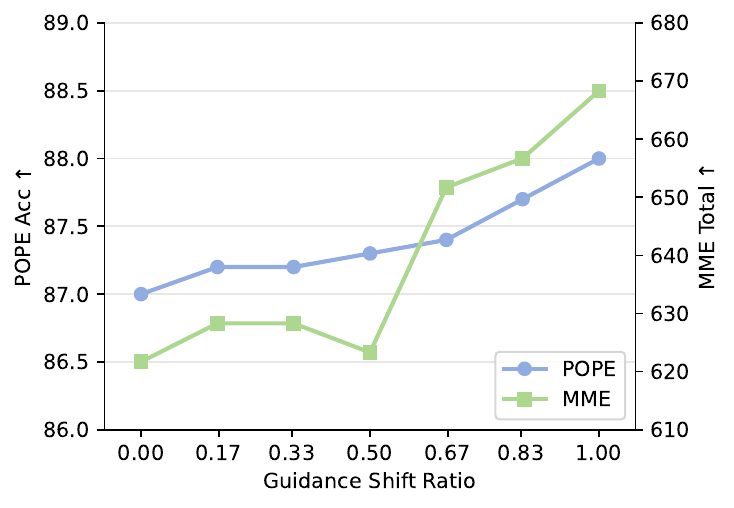}
        \vspace{2pt}
        \centerline{\scriptsize  (A)}
    \end{minipage}\hfill
    \begin{minipage}{0.33\linewidth}
        \centering
        \includegraphics[width=\linewidth]{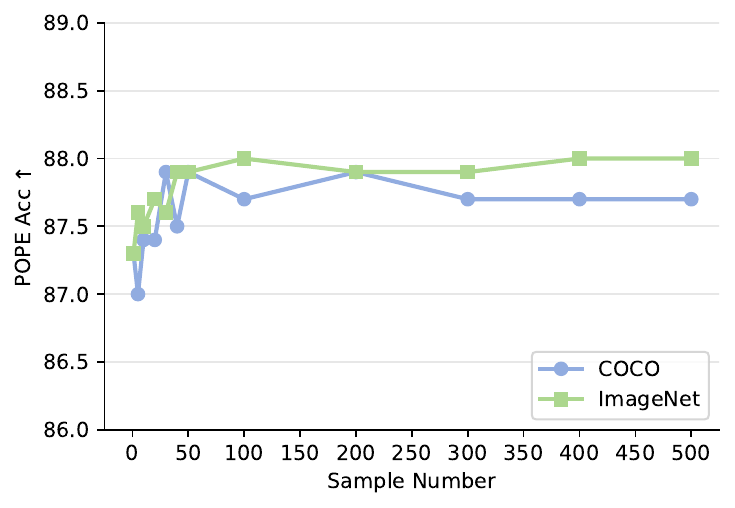}
        \vspace{2pt}
        \centerline{\scriptsize  (B)}
    \end{minipage}\hfill
    \begin{minipage}{0.33\linewidth}
        \centering
        \includegraphics[width=\linewidth]{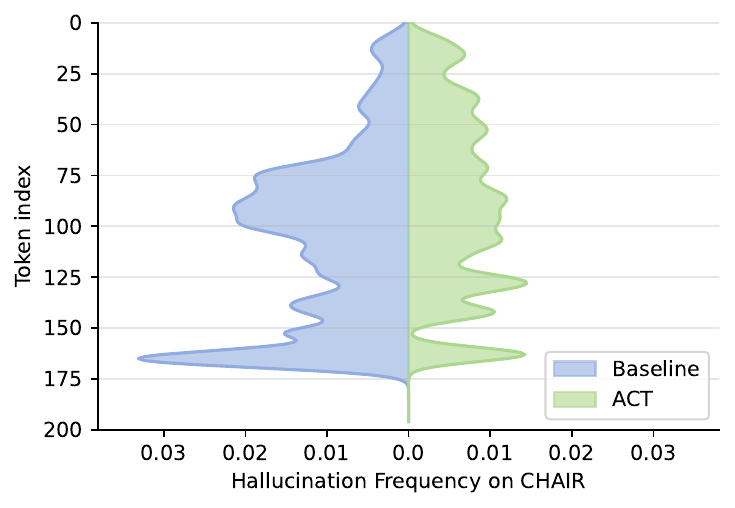}
        \vspace{2pt}
        \centerline{\scriptsize  (C)}
    \end{minipage}
    
    \caption{Visualizations of ablation study results. (A) Impact of the guidance shift between dynamic and static heads. (B) Robustness to calibration set size and source domain. (C) Positional distribution of hallucinations in CHAIR evaluation.}
    \label{fig:ablation}
\end{figure}

\textbf{Dynamic vs. Static Head Enhancement.}
We further investigate the impact of shifting the guidance focus between static and dynamic heads. We introduce a \textit{Guidance Shift Ratio} to control the enhancement scale distribution. A ratio of 1.0 allocates the full scale exclusively to dynamic heads (our default setting), and 0.0 allocates it entirely to static heads. Intermediate values proportionally interpolate the guidance between these subsets. As shown in Fig.~\ref{fig:ablation}(A), increasing the ratio toward dynamic heads consistently improves performance on the POPE and MME benchmarks. The peak performance is observed when the ratio reaches 1.0, confirming that dynamic heads are the most effective intervention targets. Prioritizing them maximizes proactive visual exploration, whereas targeting static heads yields suboptimal results. This contrast further validates the accuracy of our selection strategy based on STCS.

\textbf{Robustness to Calibration Set Selection.}
To assess the sensitivity of STCS-based head profiling, we varied both calibration sample size and source domain. As shown in Fig.~\ref{fig:ablation}(B), ACT performance on POPE remains remarkably stable once a minimal statistical threshold is met. Crucially, ACT consistently surpasses the baseline regardless of whether the calibration data is in-domain (MSCOCO~\cite{mscoco}) or out-of-domain (ImageNet~\cite{deng2009imagenet}), even with as few as 50 images. These results suggest that the identified dynamic heads capture intrinsic, generalizable visual processing patterns within the LVLM rather than dataset-specific biases. This robustness ensures the plug-and-play applicability of ACT across diverse scenarios without the need for extensive in-domain data.

\textbf{Comparison with Other Decoding Strategies.}
Because the SCA module aggregates visual evidence across parallel textual contexts, it functions similarly to traditional decoding methods like beam search. We compare it against standard decoding strategies on the CHAIR benchmark. As Table~\ref{tab:ablation_decoding} shows, SCA achieves the lowest CHAIR scores while preserving a highly competitive F1 score. This success likely occurs because SCA effectively smooths the visual attention variance induced by linguistic uncertainty.

\textbf{Mitigation of Long-Term Hallucination.}
To better understand the temporal impact of our mitigation strategy, we evaluate the distribution of hallucinated tokens across different decoding steps on the CHAIR benchmark. Fig.~\ref{fig:ablation}(C) compares the average hallucination probability at each token index for the baseline and ACT. As sequence length surpasses index 75, baseline hallucination probability noticeably increases due to the accumulation of linguistic uncertainty. In contrast, ACT significantly suppresses these delayed hallucinations. This confirms that our adaptive context integration effectively mitigates cascading errors and maintains factual grounding throughout extended sequences.

\begin{figure}[t]
    \centering
    \begin{minipage}{0.49\linewidth}
        \centering
        \includegraphics[width=\linewidth]{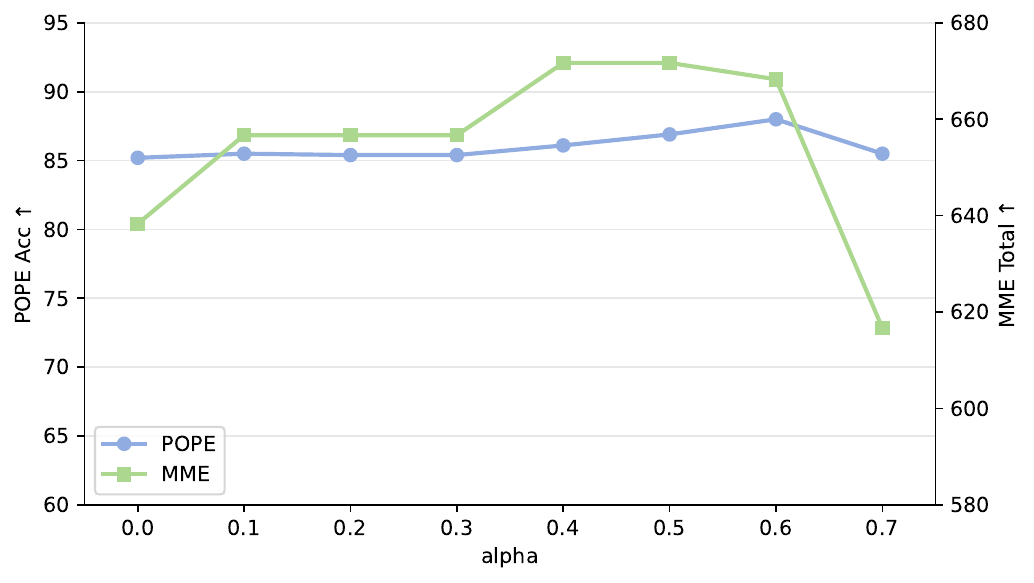}
        \vspace{-5pt}
    \end{minipage}\hfill
    \begin{minipage}{0.49\linewidth}
        \centering
        \includegraphics[width=\linewidth]{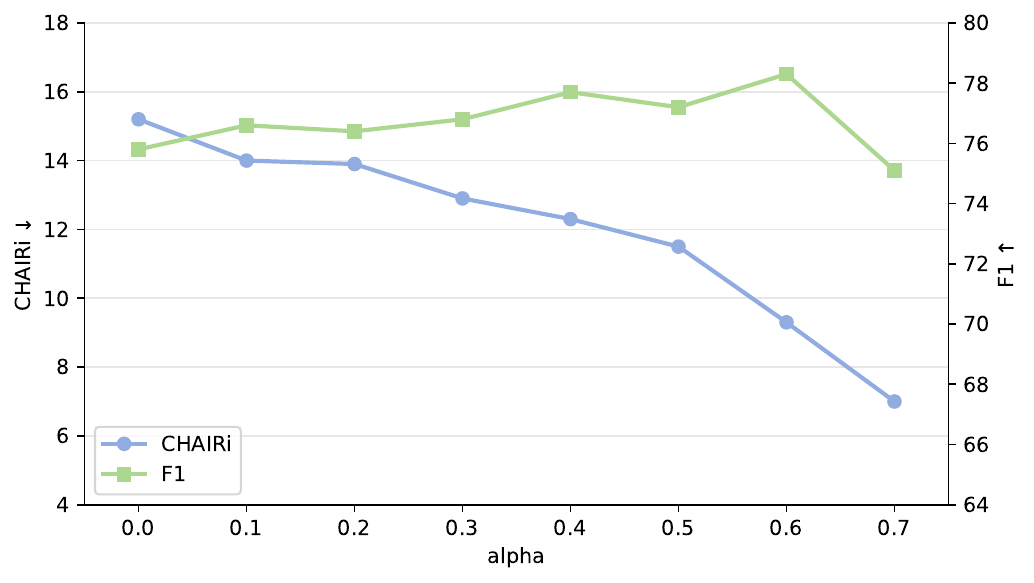}
        \vspace{-5pt}
    \end{minipage}
    \caption{Effect of the guidance scale of ACT on discriminative and generative tasks.}
    \label{fig:ablation2}
\end{figure}

\textbf{Effect of Guidance Scale.} We further investigate of ACT to the guidance scale $\alpha$ across both discriminative and generative metrics. As illustrated in Figure~\ref{fig:ablation2}, increasing $\alpha$ from 0.0 yields steady performance gains, indicating that appropriate scaling effectively prioritizes visual evidence over misleading linguistic priors. However, we observe a sharp performance collapse when $\alpha$ exceeds the threshold of 0.6. This degradation suggests that excessive guidance intensity overpowers the model's inherent linguistic coherence, leading to distorted text generation. Since $\alpha$=0.6 achieves the optimal balance between visual grounding and linguistic fluency, we adopt it as the default setting for ACT.

\subsection{Qualitative Results}

\begin{figure}[t]
\centering
\includegraphics[width=\linewidth]{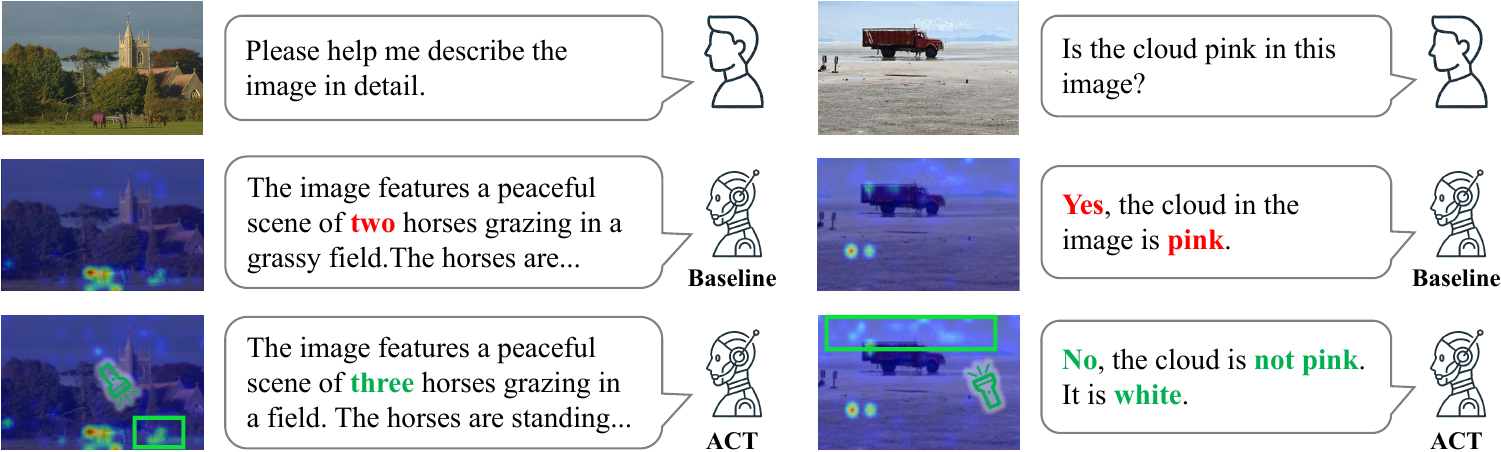}
\caption{Qualitative comparison of visual attention maps between the baseline and our proposed ACT on LLaVA-1.5-7B.}
\label{fig:qualitative_vis}
\end{figure}

To provide intuitive insights into the efficacy of our proposed approach, we visualize the attention maps generated by the ACT-enhanced model compared to the baseline. As illustrated in Fig.~\ref{fig:qualitative_vis}, the qualitative comparisons reveal distinct attentional behaviors across sample inputs.

While the baseline model often exhibits restricted or misaligned focus, ACT proactively explores a broader visual context while maintaining precise attention on critical objects. This dual improvement in breadth and accuracy ensures factual grounding, directly driving the quantitative reduction in hallucinations.

\section{Limitations}
While ACT effectively mitigates hallucinations in LVLMs, it introduces specific trade-offs in efficiency and output characteristics. Architecturally, the parallel contextual branches within the SCA module incur additional memory overhead and a slight increase in inference latency. Furthermore, by enforcing strict grounding of visual evidence, the model adopts a more cautious generation strategy, which can occasionally lead to reduced output length and a lower recall of fine-grained details. Finally, while ACT excels at resolving object and attribute-level inconsistencies, addressing complex multi-step reasoning or relational hallucinations remains an open challenge for future work.

\section{Conclusion}
In this paper, we introduced the Adaptive Context inTegration (ACT) method to tackle the pervasive issue of visual hallucinations in LVLMs. ACT pioneers a fundamental shift from reactive alignment to proactive context anticipation by seamlessly unifying two core modules: Visual Context Exploration (VCE) and Semantic Context Aggregation (SCA). Specifically, VCE operates on the visual modality, utilizing a spatio-temporal profiling strategy to encourage the model to explore dynamic visual contexts by amplifying targeted dynamic attention heads. Concurrently, SCA operates during text generation, marginalizing parallel contextual hypotheses to effectively smooth the attention variance caused by linguistic uncertainty. Extensive evaluations across multiple state-of-the-art LVLM architectures demonstrate that ACT, as a training-free and plug-and-play intervention, significantly mitigates visual hallucinations while preserving the models' fundamental generation capabilities.

\clearpage  
\bibliographystyle{splncs04}
\bibliography{main}

\clearpage

\end{document}